\documentclass{sig-alternate}
\usepackage{url}
\usepackage{amsfonts}
\usepackage{amsmath}
\usepackage{amssymb}
\usepackage{graphicx}
\usepackage{color}
\usepackage{float}
\usepackage{caption}
\usepackage{subcaption}
\DeclareCaptionType{copyrightbox}

\newcommand{\cphydra}{CPHydra}

\newcommand{\satzilla}{SATzilla}
\newcommand{\avg}{\text{avg}}  
\newcommand{\cv}{\text{CV}} 
\newcommand{\ent}{\text{H}} 
\newcommand{\flatzinc}{FlatZinc}

\newcommand{\var}{\mathcal{X}}
\newcommand{\dom}{\mathcal{D}}
\newcommand{\con}{\mathcal{C}}
\newcommand{\csp}{\mathcal{P = (X, D, C)}}

\newcommand{\tool}{\texttt{mzn2feat}}
\newcommand{\XCSPZinc}{\texttt{xcsp2mzn}}

\makeatletter
\def\url@leostyle{%
  \@ifundefined{selectfont}{\def\UrlFont{\sf}}{\def\UrlFont{\small\ttfamily}}}
\makeatother
\begin{document}
%

\title{An Enhanced Features Extractor \\ for a Portfolio of Constraint 
Solvers}
\subtitle{(Long Version)}

\numberofauthors{3}
\author{
%
%
\alignauthor
Roberto Amadini\\
       \affaddr{University of Bologna/INRIA}\\
\alignauthor Maurizio Gabbrielli\\
       \affaddr{University of Bologna/INRIA}\\
\alignauthor Jacopo Mauro\\
       \affaddr{University of Bologna/INRIA}\\
}

\maketitle
\begin{abstract}
Recent research has shown that a single arbitrarily efficient solver can 
be significantly outperformed by a \emph{portfolio} of possibly slower on-average solvers.
The solver selection is usually done by means of (un)supervised learning techniques which 
exploit features extracted from the problem specification.
In this paper we present an useful and flexible framework that is able to extract an 
extensive set of features from a Constraint (Satisfaction/Optimization) Problem defined 
in possibly different modeling languages: MiniZinc, FlatZinc or XCSP. 
We also report some empirical results showing that the performances that can be obtained using 
these features are effective and competitive with state of the art CSP portfolio techniques.
\end{abstract}

%
%

\section{Introduction}
\label{sec:introduction}
The past decade has witnessed a significant increase in the number
of constraint solving  systems deployed for solving 
Constraint Satisfaction Problems (CSPs).
It is well recognized within the field of Constraint Programming that
different solvers are better when solving different problem instances,
even within the same problem class~\cite{DBLP:journals/ai/GomesS01}.
It has also been shown in other areas, such as SATisfiability
testing~\cite{DBLP:conf/cp/XuHHL07} and
Integer Linear Programming~\cite{DBLP:conf/cp/Leyton-BrownNS02},
that the best on-average solver can be
out performed by a \emph{portfolio} of possibly slower on-average solvers. 
In a nutshell, a portfolio approach \cite{DBLP:journals/ai/GomesS01} for constraint solving can be 
seen as a methodology that exploits the significant variety in performances observed in different 
algorithms and combines them in order to create a globally better solver.
A crucial step for the performance of a portfolio solver is the selection of (one of) the solvers composing 
the portfolio for solving a specific problem instance.
Such a selection process is usually performed by using Machine Learning techniques
based on \emph{features} extracted from the instances that need to be solved.

Portfolio approaches have been extensively studied and used in the SAT solving field.
The large number of different SAT solvers available, the presence of 
a standard input language and the availability of a huge dataset of instances has fostered the study of 
how different solvers can be exploited in order to improve performances, thus bringing to the definition of several 
portfolio solvers for SAT.

Unfortunately, no such a growth exists in the CSP field, where the only solver which uses a portfolio approach 
is the Case-Based Reasoning system CPHydra~\cite{cphydra}. There are several reasons for this gap between CSP and SAT. 
First of all, the CSP solving field is more complex than SAT: constraints can be arbitrary complex (e.g. 
global constraints like \emph{regular} or \emph{bin-packing}) and some of them are supported by only a few solvers.
Moreover, no standard input language for CSP exists and there are no immediately available big dataset 
for constraint solving problems. These limitations affect also CPHydra: indeed, it can only treat problems 
expressed in the XCSP format, it uses a rather small portfolio (just 3 solvers) and it can extract only a limited 
number of features from a CSP model (a set of 36 features extracted by Mistral solver~\cite{mistral_old}). 
Nevertheless, \cphydra~was able to win the 2008 International CSP Solver Competition: this witnesses that portfolios 
may be powerful also in CSP domain.


In this paper we address these problems and we perform a first step in the direction of filling the 
gap between SAT and CSP portfolio solvers.
In particular, we present a framework that is able to:
\begin{itemize}
 \item extract an exhaustive set of 155 features from a general MiniZinc~\cite{Nethercote07minizinc:towards} specification;
 \item process different formats like XCSP and FlatZinc \\ through a simple pre-processing phase;
 \item deal with both satisfaction and optimization problems;
 \item provide a fundamental support for a portfolio of constraint solvers.
\end{itemize}

We decided to use MiniZinc as source format of our tool because it is 
nowadays the most used, supported and general language to specify constraint problems. 
MiniZinc supports also optimization problems and is the 
source format used in the MiniZinc challenge~\cite{DBLP:journals/constraints/StuckeyBF10}, the only 
surviving international 
competition to evaluate the performances of constraint solvers.
From a technical point of view, MiniZinc is compiled into the low 
level language FlatZinc using \emph{ad hoc} global constraint redefinitions and this format is then 
used to extract the features.\footnote{We decided to start from a MiniZinc specification in order 
to capture an extensive set of global constraints that may be lost during the 
compilation in FlatZinc, depending on the solver specific redefinitions.}
However, our framework offers a full compatibility with XCSP and FlatZinc. Indeed, on one hand, 
we developed a compiler 
\XCSPZinc\ for converting problem instances from XCSP to MiniZinc by preserving the most important 
global constraints. 
On the other hand, the feature extractor tool called \tool\ supports natively the FlatZinc format 
and can extract the features from every FlatZinc model (possibly ignoring unknown solver specific redefinitions).

For a preliminary validation of the framework, following the 
approach presented in \cite{DBLP:conf/cpaior/AmadiniGM13} we built different portfolio 
solvers consisting of up to 11 solvers taken from those used in MiniZinc challenge 2012. 
We used off-the-shelf machine learning algorithms as well as state of the art portfolio 
techniques in order to exploit the new extracted features.
To obtain an extensive evaluation as possible the tests were conducted by using a dataset obtained 
by combining the CSP instances of the MiniZinc benchmark with the dataset of the last two 
International Constraint Solver Competitions (ICSCs) \cite{CSC09}. 
Results indicate that the performances that can be obtained using the new set of features are competitive 
with state of the art CSP portfolios techniques.

\emph{Paper structure.} 
In Section~\ref{sec:prelim} we recall some preliminary notions and we discuss the related 
literature.
In Section~\ref{sec:tools} we describe the technical details of our framework 
while in 
Section~\ref{sec:validation}
we discuss the empirical validation. Finally, in Section~\ref{sec:conclusion} we report some 
concluding remarks.

\section{Background}
\label{sec:prelim}
In this Section we introduce some preliminary notions that we need later in the paper, 
then we discuss the related work.
\subsection{Preliminaries}
A \textit{Constraint Satisfaction Problem} (CSP) $\csp$ consists of a finite set of
va\-ria\-bles $\var$, each of which associated with a domain $D_x \in \dom$ of possible 
values that a variable $x$ could take, and a set of constraints $\con$ that define the set
of allowed assignments of values to the variables~\cite{DBLP:journals/ai/Mackworth77}.
Given a CSP the goal is normally to find a solution, that is an assignment to the
variables that satisfies all the constraints of the problem, through one suitable constraint
solver.

\textit{Machine Learning} (ML) is a broad field that uses concepts from computer science, mathematics,
statistics, information theory, complexity theory, biology and cognitive 
science~\cite{Mitchell97} to ``construct computer programs that automatically improve with experience''.
In particular, \emph{classification} is a well-known ML problem that, given a finite number of classes (or categories), 
consists in identifying to which class belongs each new observation. This problem is solved 
by using an appropriate classifier which is essentially a function mapping a new instance - characterized by one or more 
discrete or continuous \emph{features} - to one class~\cite{Mitchell97}. The classifier is defined on the 
basis of a dataset of instances whose class is already known, trying to exploit such a knowledge to properly 
classify each new instance. 

As previously mentioned a \emph{portfolio} approach~\cite{DBLP:journals/ai/GomesS01} is a general methodology that, 
in our case, allows to exploits the synergies of different (constraint) solving algorithms in order to obtain a globally 
better solver. We can then consider a portfolio of solvers as a particular solver $S$ 
consisting of a number $m > 1$ of different (constituent) solvers $S_1, \dots, S_m$. 
When a new problem $P$ is given, the portfolio tries to predict which is the best constituent solver $S_i$ for solving 
the specific problem $P$ and then uses it for solving $P$. This solver selection process, which is clearly a fundamental part 
for the success of the approach, is usually performed by using ML techniques. In particular, classification techniques are 
often used to make predictions on the basis of the features extracted from a relevant set of problem instances. 
There are also hybrid approaches that integrate ML with other techniques (e.g. Integer Programming) in order to improve 
the accuracy of the predictions and to maximize the number of solved problems.

\subsection{Related Work}
\label{subsec:rel_work}
To the best of our knowledge, \cphydra~\cite{cphydra} is currently the only CSP solver which uses a 
portfolio approach. 
For the feature extraction it uses the code of Mistral, one of its constituent solvers, that is 
able to extract only 36 features. In \cite{KMMO11} the feature extraction code was improved 
allowing the extraction of few additional features. 
The main weakness of \cphydra\ concerns the fact that it is not very scalable w.r.t. the number 
of the constituent solvers since it has to solve an NP-hard problem to decide the schedule of 
solvers to use to solve an instance. Moreover, it assumes that problem instances are formulated in 
the XCSP format, which is less expressive (and less used today) than MiniZinc.

On the other hand, there are several portfolio solvers for SAT.
3S~\cite{DBLP:conf/cp/KadiogluMSSS11} is a SAT solver that conjugates a fixed-time static solver 
schedule with the dynamic selection of one long-running component solver.
3S solves the scalability issues of \cphydra\ because the scheduling is computed offline and covers only 10\% 
of the time limit. If a given instance is not yet solved after the short runs, a designated solver 
is chosen at runtime (using a $k$-nearest neighbors algorithm) and executed.

\satzilla~\cite{DBLP:conf/cp/XuHHL07} is a SAT solver that relies on runtime prediction models to 
select the solver that (hopefully) has the fastest running time on a given problem instance.
Its last version~\cite{DBLP:conf/sat/XuHHL12}, which consistently outperforms the previous ones, uses a weighted 
random forest approach provided with an explicit cost-sensitive loss function punishing misclassifications in 
direct proportion to their impact on portfolio performance.

In \cite{DBLP:conf/cpaior/MalitskyS12} the Instance-Specific Algorithm Configuration 
tool ISAC~\cite{DBLP:conf/ecai/KadiogluMST10} has been used as solver selector. The aim of ISAC is to 
optimally tune the solver parameters on the basis of the given instance features, behind the 
primary assumption that a solver will have consistent performance on instances that are clustered together.

In \cite{DBLP:conf/cpaior/AmadiniGM13} an empirical evaluation and comparison of 
 portfolio approaches is presented. 
Different portfolio sizes and evaluation metrics were used on a dataset of XCSP instances taken from 
the last two ICSCs.

Recalling that in this work we focus only on sequential approaches, we would however mention some porfolio-based 
parallel SAT solvers, like ManySAT~\cite{Hamadi09manysat:a}, PeneLoPe~\cite{AHJ+-12-3} and ppfolio~\cite{ppfolio_web}.

Other recent works show that the interest in algorithm runtime prediction is quite general and growing. 
A detailed overview of the state of the art in this context is provided in 
\cite{DBLP:journals/corr/abs-1211-0906} that also describes new features for predicting 
algorithm runtime for SAT, MIP (Mixed Integer Programming), and TSP (Traveling Salesperson) problems.

In \cite{Arbelaez09onlineheuristic}, \cite{DBLP:conf/ictai/ArbelaezHS10} ML techniques are used to enhance the performances 
of a single CSP solver by dynamically adapting its search heuristics. This work lists an extensive 
set of features to train and improve the heuristics model through Support Vector Machines. 

Feature filtering techniques for ISAC tool are described in \cite{citeulike:10395539}: instead of using traditional approaches, 
the authors introduce new evaluation functions 
to quickly evaluate subsets of features. 
Numerical results on both SAT and CSP domains show that the number of features can be significantly 
reduced while often providing considerable performances gains. Moreover, in \cite{snnap} the authors introduce SNNAP 
(Solver-based Nearest Neighbors for Algorithm Port\-fo\-lios), an alternative view of ISAC which uses the existing features to 
predict the best three solvers for a particular instance.
A brand new classifier that selects solvers based on a
Cost-Sensitive Hierarchical Clustering (CSHC) model is presented in \cite{DBLP:conf/ijcai/MalitskySSS13}. 
CSHC solver won 2 gold medals in SAT competition 2013.

Finally, a number of tools are being developed in order to improve portfolio solvers usability. 
\texttt{snappy} (Simple Neigh\-bor\-hood-based Algorithm Portfolio in PYthon) \cite{DBLP:conf/sat/SamulowitzRSS13} is a 
simple and training-less algorithm portfolio which relies on a nearest neighbors prediction mechanism. 
LLAMA (Leveraging Learning to Automatically Manage Algorithm) \cite{kotthoff_llama_2013} is instead a framework that facilitates 
the exploration of different portfolio techniques on any problem domain, by supporting the most common solver selectors and  
possibly combining them.

\section{Framework}
\label{sec:tools}
In this section we present the technical details of our framework, introducing the compiler 
\XCSPZinc\ and the features extractor \tool\ (together with a detailed list of the features it 
extracts).


\subsection{\XCSPZinc\ and \tool\ }
MiniZinc is nowadays the most used language to encode CSPs while XCSP was mainly used in 
the past for the International Constraint Solver Competition (ICSC), which ended in 2009. Nevertheless, the ICSC dataset 
is by far the biggest dataset of CSP instances existing today. Hence, in order to exploit such a dataset
for building better portfolios, we developed a compiler from XCSP to MiniZinc.

\XCSPZinc\ was developed by adapting \texttt{x4g}~\cite{DBLP:conf/cilc/MoraraMG11}, a converter from 
XCSP to Gecode~\cite{gecode_fzn} used 
in particular to support the XCSP abridged notation. Since we focused mainly on CSP we did not consider XCSP extensions 
like weighted constraints or quantifiers over constraints. All the code is written in C++ using the well known 
\texttt{libxml2} libraries. 

Exploiting the fact that MiniZinc is more expressive than XCSP (i.e. the majority of the 
primitive constraint of XCSP are also primitive constraints of MiniZinc) the translation was 
straightforward. The only notable difference was the compilation of extensional constraints (i.e. 
relations explicitly expressed in terms of all the allowed or not allowed tuples) which are a 
native feature in XCSP only.
To overcome this limitation we used the \emph{table} global constraint for encoding the allowed set of 
tuples and a conjunction of disjunctions of inequalities  for mapping the forbidden set of tuples.

As far as global constraints are concerned, XCSP supports the majority of the global constraints 
defined in the Global Constraint Catalog~\cite{Beldiceanu:2007:GCC:1232658.1232664}. Since in this catalog 
there are hundreds of global constraints, a full XCSP support means to provide an encoding for a huge 
number of them. However, this is out of our scope and we have chosen to 
support only the subset of the global constraints used in the ICSC.


\tool\ is a tool that allows to extract from a MiniZinc 
model a set of 155 features: 144 are static features and are obtained by parsing the problem instance, while 11 are dynamic 
and are obtained by running the Gecode solver for a short run (for a detailed description of the features 
please see Section \ref{subsec:features}).
Since the complexity of the MiniZinc language (in particular the possibility of using control 
flow statements) makes the extraction of the syntactical features quite difficult, we decided to not process directly the MiniZinc 
instances. We instead compile them to FlatZinc~\cite{mzn2fzn}, a lower level language 
having a syntax that is mostly a subset of MiniZinc and that can be obtained from 
MiniZinc by using the \texttt{mzn2fzn} tool provided by the MiniZinc suite.

To develop \tool\ we first generated the FlatZinc parser of the MiniZinc suite 
by using the standard Flex and Bison  parser tools. Then, the generated parser was extended by integrating 
suitable C++ code for extracting the static features (for example the number of constraints, their arity, etc.). 

The compilation to FlatZinc raised some design choices, since global 
constraints defined in MiniZinc can be translated in different ways. For example, the 
\texttt{alldifferent} global constraint is decomposed by default into 
 a conjunction of inequalities. However, if for instance the target solver of the compilation is 
Gecode, specific definitions can be used to avoid its decomposition.
This is a key feature of MiniZinc: starting from a general model each solver can produce a 
specialized FlatZinc by redefining the global specifications. \\
Since a proper treatment of global constraints can dramatically 
improve the solver performances, we thought that keeping track of how and what global 
constraints are used is rather important. For this reason we decided to consider as input format the FlatZinc 
obtained 
by using Gecode redefinitions.
 Such a choice is justified by the fact that Gecode won the gold medal in all categories of the 
 MiniZinc Challenge 2012~\cite{mznc2012} and it handles natively 47 different global 
constraints.
 
The specific FlatZinc model obtained in this way is also exploited to collect the dynamic 
features. This was done by launching Gecode interpreter \texttt{fz} for short runs (2 
seconds).
 
Summarizing, given a generic MiniZinc model $M$ in input, \tool\ does the following: 
\begin{enumerate}
 \item it translates $M$ into the corresponding FlatZinc $F_M$ specification by using Gecode global redefinitions; 
 \item it extracts static features from $F_M$ by using a suitable parser;
 \item it extracts dynamic features from $F_M$ by running the \texttt{fz} interpreter of Gecode for 2 seconds.
\end{enumerate}
We remark that step 2) is applicable to every FlatZinc model $F$ (possibly ignoring the unknown solver-specific 
redefinitions). Moreover, steps 2) and 3) are totally independent and therefore could be parallelized or even reversed. 
For instance, it could be useless to compute 
the static features if the given instance is solved by Gecode while trying to compute the dynamic 
features.

\subsection{Features description}
\label{subsec:features}
In this section we present a detailed list of all the 155 numeric features extracted by \tool.
We tried to collect a set of features as exhaustive and general as possible, taking inspiration from and adapting
those presented in \cite{DBLP:journals/corr/abs-1211-0906}, \cite{DBLP:conf/ictai/ArbelaezHS10}. 
Although some of these features are quite generic (e.g., the number of 
variables or constraints) others are specific to FlatZinc (e.g. search annotations) or to Gecode 
(the global constraints features). For more details about these technical details we defer the 
interested reader to \cite{mzn2fzn}, \cite{fzn_spec}, \cite{gecode_fzn}.

In the following we denote by $\var$ the set of the \emph{unbounded} variables and, if not specified, 
with the term ``variable'' we refer to unbounded variables.
A variable for us is bounded  if it is either bounded to a constant value (and in this case it is called {\em constant} for short) 
or it is bounded to another variable (i.e, it is an \emph{alias}). If $x_1, \dots, x_k$ 
are aliases of a variable $x$,  we compute the corresponding features by assuming to replace each $x_i$ by  
the aliased variable $x$.  Moreover, we denote by $dom(x)$ the domain size of a variable $x \in \var$, by $deg(x)$ its degree 
(i.e. the number of constraints $c \in \con$ in which $x$ occurs) and we define $\overline{\var} \subseteq \var$ as 
$\overline{\var} = \{ x  \in \var : deg(x) \neq 0 \}$.

Similarly, we will denote by  $\con$ the set of constraints that constrain at least one 
variable. For each $c \in \con$, we denote by  $Var(c)$ the set of variables that occur in $c$, by $deg(c) 
= |Var(c)|$, $dom(c) = \log{(\prod_{x \in Var(c)}{dom(x)})}$ (we use the logarithm since for large domains 
the computation of $\prod_{x \in Var(c)}{dom(x)}$ may cause an overflow), and $ari(c)$ 
 the arity of $c$ (i.e. the number of its arguments; note that $deg(c) = ari(c)$ iff all the variables 
 occurring in $c$ are distinct).
 
Finally, we will denote respectively by $\min$, $\max$, $\avg$, $\cv$, and $\ent$ 
the minimum, maximum, average, variation coefficient and entropy values.

\subsubsection*{Static Features}
We extracted 144 static features grouped in the following different categories.

\begin{itemize}
  \item \textbf{Variables} (27): 
 \begin{itemize}
  \item the number of variables $|\var|$, the number $cv$ of constants, the number $av$ of aliases, 
  the ratio $\frac{av + cv}{|\var|}$, the ratio $\frac{|\var|}{|\con|}$;
  \item the number of \emph{defined} variables (i.e. defined as a function of other variables), and the number of \emph{introduced} 
variables (i.e.  auxiliary variables introduced during the \flatzinc\ conversion);
  \item $\log{(\prod_{x \in \var}{dom(x)})}$, $\log{(\prod_{x \in \overline{\var}}{deg(x)})}$;
  \item $\sum_{x \in \var}{dom(x)}$, $\sum_{x \in \var}{deg(x)}$,
  $\sum_{x \in \overline{\var}}{\frac{dom(x)}{deg(x)}}$;
  \item $\min$, $\max$, $\avg$, $\cv$, and $\ent$ of $\{dom(x): x \in \var\}$;
  \item $\min$, $\max$, $\avg$, $\cv$, and $\ent$ of $\{deg(x): x \in \var\}$;
  \item $\min$, $\max$, $\avg$, $\cv$, and $\ent$ of 
  $\{\frac{dom(x)}{deg(x)}: x \in \overline{\var} \}$.
 \end{itemize}
 
 \item \textbf{Domains} (18): Since variables could have different domains, we compute the number 
of:
 \begin{itemize}
  \item boolean variables $bv$ and the ratio $\frac{bv}{|\var|}$;
  \item float variables $fv$ and the ratio $\frac{fv}{|\var|}$;
  \item integer variables $iv$ and the ratio $\frac{iv}{|\var|}$;
  \item set variables $sv$ and the ratio $\frac{sv}{|\var|}$.
 \end{itemize}
 Moreover, we compute the number of:
 \begin{itemize}
  \item array constraints $ac$ and the ratio $\frac{ac}{|\con|}$;
  \item boolean constraints $bc$ and the ratio $\frac{bc}{|\con|}$;
  \item int constraints $ic$ and the ratio $\frac{ic}{|\con|}$;
  \item float constraints $fc$ and the ratio $\frac{fc}{|\con|}$;
  \item set constraints $sc$ and the ratio $\frac{sc}{|\con|}$.
 \end{itemize}
 
 \item \textbf{Constraints} (27):
 \begin{itemize}
  \item the total number of constraints $|\con|$, the ratio $\frac{|\con|}{|\var|}$, the total number of 
constraints using \texttt{boundsZ} (or \texttt{bounds}), \texttt{boundsR}, \texttt{boundsD}, 
\texttt{domain} or \texttt{priority} specific FlatZinc annotations;
  \item $\log{(\prod_{c \in \con}{dom(c)})}$, $\log{(\prod_{c \in \con}{deg(c)})}$;
  \item $\sum_{c \in \con}{dom(c)}$, $\sum_{c \in \con}{ari(c)}$,
   $\sum_{c \in \con}{\frac{dom(c)}{deg(c)}}$;
  \item $\min$, $\max$, $\avg$, $\cv$, and $\ent$ of $\{dom(c): c \in \con\}$;
  \item $\min$, $\max$, $\avg$, $\cv$, and $\ent$ of $\{deg(c): c \in \con\}$;
  \item $\min$, $\max$, $\avg$, $\cv$, and $\ent$ of 
  $\{\frac{dom(c)}{deg(c)}: c \in \con\}$.
 \end{itemize}
 
 \item \textbf{Global Constraints} (29): We consider the total number $gc$ of global constraints, 
the ratio $\frac{gc}{|\con|}$ and the number of global constraints for each one of the  27 
equivalence classes in which we have grouped the 47 global constraints that Gecode supports.\footnote{
As an example, the class \emph{bool lin} includes the constraints 
\textit{bool\_lin\_eq, bool\_lin\_ne, bool\_lin\_le, bool\_lin\_lt, bool\_lin\_ge, bool\_lin\_gt}.}

 \item \textbf{Graphs} (20): In order to capture the interactions between variables and 
 constraints we computed the followings non oriented graphs:\footnote{Constraint Graph and Variable Graph
 are also known as \emph{Dual} and \emph{Primal} graph respectively.}
 \begin{itemize}
 \item \textit{Constraint Graph} (CG): the graph obtained add\-ing a node for each constraint $c \in 
\con$ and an edge between 
$c_1$ and $c_2$ iff they share at least one variable (i.e. $Var(c_1) \cap Var(c_2) \neq \emptyset$);
  \item \textit{Variable Graph} (VG): the graph obtained adding a node for each variable $x \in 
\var$ and an edge between 
$x_1$ and $x_2$ iff they occur together in at least one constraint (i.e. $\exists c \in \con. \ x_1, x_2 \in 
Var(c)$). 
 \end{itemize}
 Then, starting from the graphs and following \cite{DBLP:journals/corr/abs-1211-0906}, we computed 
 $\min$, $\max$, $\avg$, $\cv$, and $\ent$ of the:
 \begin{itemize}
  \item CG nodes degree;
  \item CG nodes clustering coefficient;
  \item VG nodes degree;
  \item VG nodes diameter, where by the diameter of a node $x$ we mean the maximum among 
the minimum distances between $x$ and each other node $y \neq x$ (we set to 0 the 
diameter of two not connected nodes)
 \end{itemize}
We noticed that for huge instances the generation of the graphs was time and space consuming. To limit the time and 
the memory needed to extract the features we have then imposed a timeout of 2 seconds to compute both CG 
and VG features. In case of timeouts these features where set to the default value of -1.
 
 \item \textbf{Solving} (11): From the solve goal we extract the following features:
 \begin{itemize}
  \item the number of \emph{labeled} variables, i.e. the variables to be assigned;
  \item goal: it can be either 1, 2, or 3 depending on the fact that the goal is  \texttt{satisfy}, \texttt{minimize}, or \texttt{maximize}, respectively;
  \item search type: the number of \texttt{bool\_search},\\ \texttt{int\_search, set\_search} 
annotations;
  \item variable choice: the number of \texttt{input order}, \\ \texttt{first\_fail}, or other 
heuristics;
  \item value choice: the number of \texttt{indomain\_min}, \\\texttt{indomain\_max}, or other 
heuristics.
 \end{itemize}
 
 \item \textbf{Objective} (12):
 Named $v$ the variable that has to be optimized, $\mu_{dom}$ and $\sigma_{dom}$ the average and the standard deviation 
of $\{dom(x) : \ x \in \var\}$ resp., $\mu_{deg}$ and $\sigma_{deg}$ the average and the standard deviation 
of $\{deg(x) : \ x \in \var\}$ resp., we compute the following features:
\begin{itemize}
 \item $dom(v)$, $\frac{dom(v)}{\mu_{dom}}$, $\frac{dom(v) - \mu_{dom}}{\sigma_{dom}}$, and $\frac{dom(v)}{deg(v)}$;
 \item $deg(v)$, $\frac{deg(v)}{\mu_{deg}}$, $\frac{deg(v) - \mu_{deg}}{\sigma_{deg}}$, and $\frac{deg(v)}{|\con|}$;
 \item the degree $de$ of $v$ in the variable graph, its diameter $di$, $\frac{de}{di}$, and $\frac{di}{de}$
\end{itemize}
 Obviously, these features make sense only for optimization problems: in case of satisfaction problems 
 (or if the denominator of the above ratios is 0) we set the default value -1 for such features.
\end{itemize}

\subsubsection*{Dynamic features}
We extracted the following 11 dynamic features:
\begin{itemize} 
 \item the number of solutions found, 
 the number $p$ of propagations performed, and 
 the ratio $\frac{p}{|C|}$;
 \item the number $e$ of nodes expanded in the search tree,
 the number $f$ of failed nodes in the search tree, and 
 the ratio $\frac{f}{e}$;
 \item the maximum depth of the search stack and 
 the peak memory allocated;
 \item the CPU time needed for converting from MiniZinc to FlatZinc,
 the CPU time required for static features computation, and
 the total CPU time needed for extracting all the features.
 \end{itemize} 

The first 8 features are collected by short runs (2 seconds) of Gecode with default 
parameters and by using \texttt{-s} 
and \texttt{-time} options for the \texttt{fz} interpreter.\footnote{
We noticed that in few cases the Gecode solving process ignores the time 
cap, probably because of a bug. In these cases we forced the interruption of \texttt{fz} after 5 seconds and set 
the default value to -1 for the 8 first dynamic features.}
 

\section{Validation}
\label{sec:validation}
Since we are not aware of tools that could extract features form MiniZinc models, making a direct 
comparison between \tool\ and other similar tools was not possible.
Therefore, although it is not the main purpose of this work to identify which are 
the most significant features for a particular portfolio approach, we decided to compare 
\tool\ with the features extractor developed by Mistral and extended in \cite{KMMO11} 
by using the ICSC dataset.
More precisely, we measured and compared the solving time and the number of problem instances solved by different portfolios 
techniques using, on one hand, the features extracted by \tool\ and, on the other hand, those extracted by Mistral.

Taking as reference the methodology and the results of \cite{DBLP:conf/cpaior/AmadiniGM13} we 
considered the most promising portfolio techniques and we evaluated their 
performances using a time limit of 1800 seconds, which is the same threshold used in ICSC. 
For building portfolios we reproduced the best performing SAT approaches of \cite{DBLP:conf/cpaior/AmadiniGM13}, 
namely, SATzilla~\cite{DBLP:conf/cp/XuHHL07} and 3S~\cite{DBLP:conf/sat/XuHHL12}, and 
the best off-the-shelf approaches, viz. 
Random Forest and SMO, by using the 
corresponding WEKA~\cite{Hall_theweka} classifiers with default parameters.

Portfolios were built by using the following 11 different solvers from the 
Mini\-Zinc Challenge 2012~\cite{mznc2012}: BProlog, Fzn2smt, CPX,
G12/FD, G12/LazyFD, G12/MIP, Gecode, izplus, MinisatID, Mistral\footnote{Note that the version of 
Mistral used in the MiniZinc challenge 2012 is  a completely new version of the solver 
and it is not the one used for extracting the features from XCSP.} and OR-Tools.
%


Every approach was tested by using a 5-repeated 5-fold cross-validation~\cite{ArlotCelisse2010}.
The dataset was randomly partitioned in 5 disjoint sets called folds. Each of these folds was 
treated in turn as the test set, considering the union of the 4 remaining folds as training data.
In order to avoid a possible \emph{overfitting} problem (i.e. a portfolio approach that adapts too well on 
the training data rather than learning and exploiting the generalized pattern) the random 
gene\-ration of the folds was repeated 5 times.
For every instance of every test set we computed the solving strategy proposed by the portfolio 
approach and we simulated it by 
checking if the solving strategy was able to solve the instance within the time cap.
We evaluated the performances of every approach in terms of Ave\-ra\-ge Solving Time 
(AST) and Percentage of Solved Instances (PSI).

In order to compare \tool~w.r.t.~the features extracted by Mistral we considered the 
instances of the ICSC successfully compiled into MiniZinc by using Gecode specifications within 900 
seconds. When the compilation time exceeds such a limit it is reasonable to assume 
that one can use directly Gecode to solve the problem instance, since recompiling the MiniZinc 
model by using the specification of a different solver would end up in wasting the entire time allowed to solve 
the instance.
We also discarded from the dataset all the instances solved by
Mistral\footnote{Here we ran the old version of Mistral, the one used in the ICSC competition.} 
and Gecode in less than 2 seconds. These instances were discarded because no 
prediction was needed: the problems were already solved during the features extraction.
In this way we end up with a benchmark consisting of 2595 instances which in the following will be called Benchmark A.

Moreover, we have also tested our tool by combining the CSP instances gathered from the 
MiniZinc benchmark and the instances of the ICSC successfully converted into MiniZinc. 
As above, we discarded from this second dataset all the instances whose compilation required more than 900 seconds 
as well as all the instances solved by Gecode in less than 2 seconds.
In this case, we ended up with a larger dataset consisting of 4642 instances: 3538 from the 
ICSC and 1104 from the Minizinc benchmark. In the following, we will refer to this dataset as 
Benchmark B.

In order to evaluate portfolio performances we simulated the execution of the solvers keeping track of the solving time 
of each solver on every CSP
and the time needed for 
the features extraction.
In average, the time needed to compute the features for the instances of Benchmark B (the larger one) was 75.61 seconds, 
with a maximum value of 897.26 seconds.
However, the median is 6.85 seconds, hence for half of 
the instances the total time required 
for computing the features was less than 7 seconds. It is worth noting that for the other half of the instances  the 
time needed to extract the features is strongly influenced by the compilation into FlatZinc 
(in average, the 47.28\% of the total extraction time).
This is due the fact that many of such instances are huge (even in the order of MB) and 
therefore their compilation is very expensive. 
Nevertheless, features extraction was performed in less than a minute for 73.52\% of the instances.

All the approaches were tested with portfolios of different sizes, considering up to a maximum of $11$ solvers.
Denoting by $\Sigma$ the set of all our solvers, for each size $n = 2, \dots, 11$ the portfolio composition was computed 
considering the subset $\Pi \subseteq \Sigma$ with cardinality $n$ which maximizes the number of solved instances 
(possible ties were broken by minimizing average solving time).


Following \cite{DBLP:conf/cpaior/AmadiniGM13} we elected MinisatID~\cite{minisatid} as backup solver, 
since it is the one that solved the greatest number of instances within the time limit of 1800 seconds.\footnote{
Thanks to the help of Broes De Cat we were able to use a debugged version of MinisatID w.r.t. the one that has competed in the last 
MiniZinc challenge.} 
The backup solver is used in case the portfolio selects a solver that fails prematurely.

%

All the code developed to conduct the experiments, together with the \XCSPZinc\ and \tool\ source 
code, is available at \cite{source_code}. 

\subsection{Test Results}
\begin{figure}[t]
\centering
\includegraphics[width=0.47\textwidth,trim=1.5cm 16cm 1.5cm 20,clip]{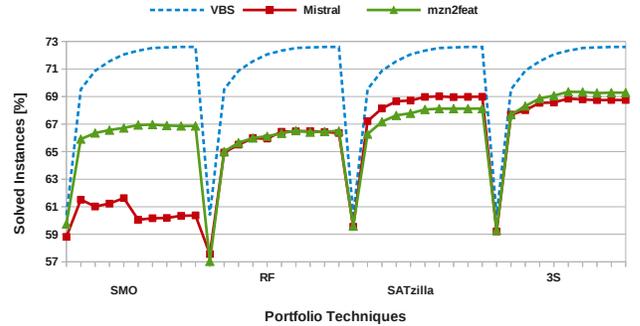}
\caption{PSI comparison using \tool\ and Mistral features on Benchmark A. For each different portfolio technique, 
the \texttt{x}-axis is sorted in ascending order according to the number of constituent solvers $n \in [2, 11]$.}
\label{fig:compare_pic}
\end{figure}
Figure \ref{fig:compare_pic} presents the results of the simulated portfolio techniques on Benchmark A by using both 
Mistral features and the features extracted by \tool.
The features for both approaches were normalized in the $[-1, 1]$ range and then used to make predictions. 
Useless features, i.e. features constants for all the instances of the dataset, were removed.
The plot presents the PSI obtained by using 4 different portfolio techniques: the first 2 are 
off-the-shelf 
approaches (Random Forest and SMO) simulated by using WEKA tool with default parameters, 
while 
the others (3S and SATzilla) comes from the SAT field. As baseline we used the Virtual Best 
Solver (VBS), 
i.e. an oracle able to choose the best solver to use for every instance.

The results clearly indicate that the performances achieved by using \tool's features are competitive 
w.r.t. those obtained by using Mistral's features.
Usually the difference is rather small even though, for instance, the SMO classifier 
allows to solve even 6.87\% of instances more by using our tool.
Considering all the tested techniques and all the portfolio sizes, in the worst case we are 
able to solve only 1.03\% less instances while, on average, 1.25\% more instances could 
be solved by using the features extracted by \tool. 
It is also worth noticing that the peak performances are reached by using 3S approach 
with the features extracted by our tool.

\begin{figure}[h]
\centering
\begin{subfigure}{1\textwidth}
\includegraphics[width=0.47\textwidth,trim=1.7cm 15cm 1cm 1cm, clip]{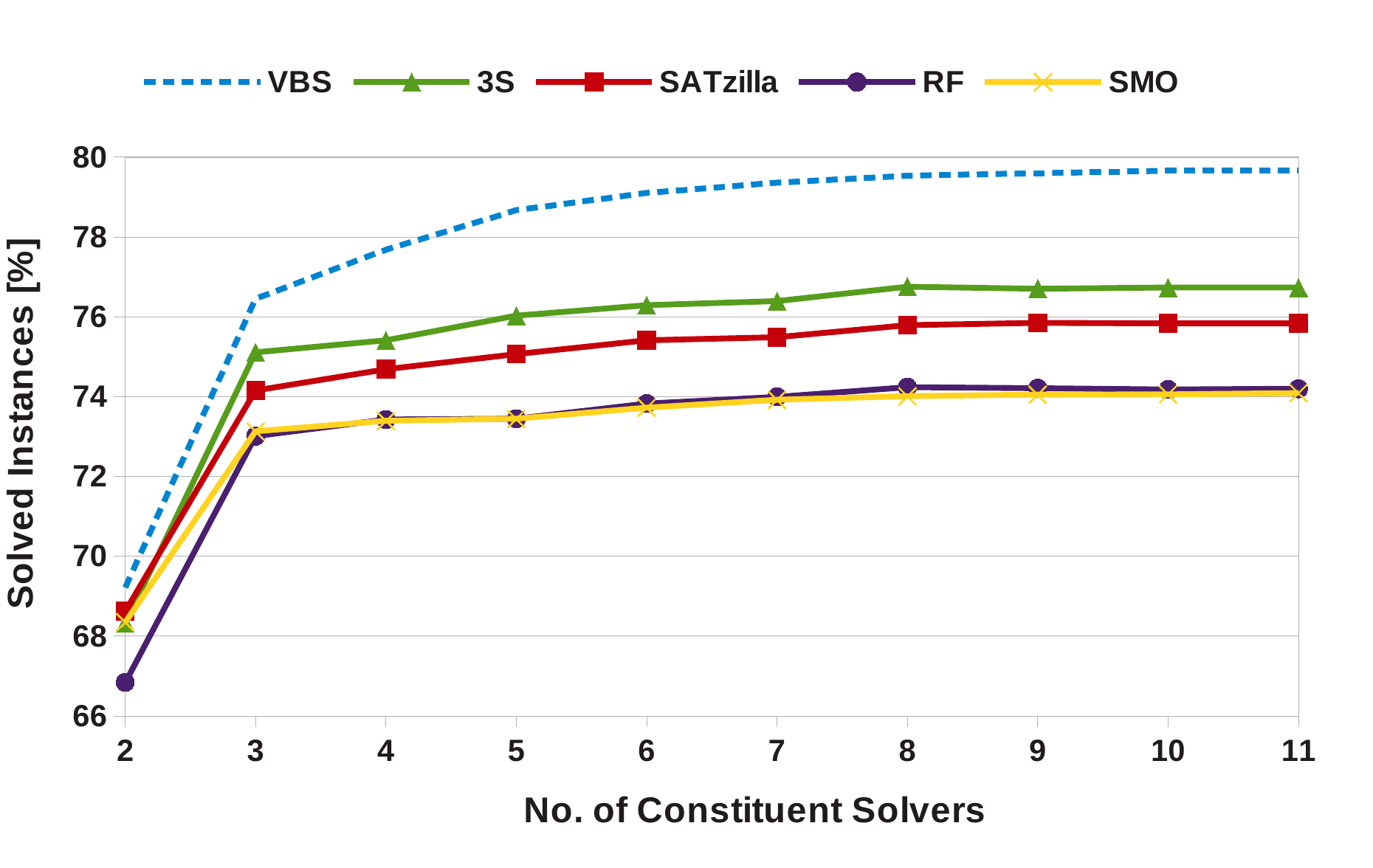}
\end{subfigure}
\begin{subfigure}{1\textwidth}
\includegraphics[width=0.47\textwidth,trim=1.8cm 15cm 1cm 2cm, clip]{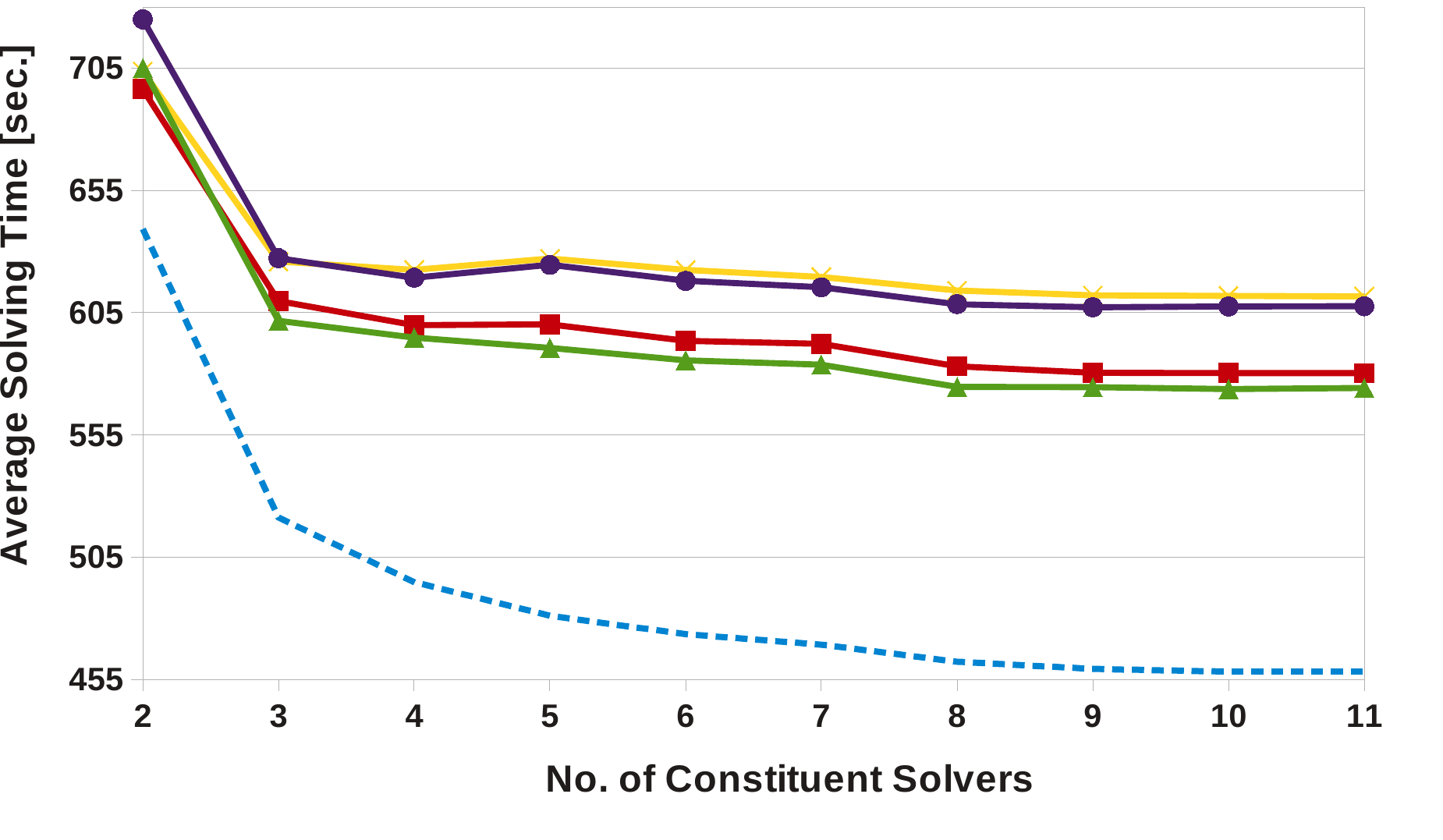}
\end{subfigure}
\caption{PSI and AST on Benchmark B.}
\label{fig:big_dataset}
\end{figure}

Figure \ref{fig:big_dataset} shows the performances achieved by using different portfolios techniques 
and the features extracted by \tool\ on the most extended Benchmark B.
The results are similar 
to those presented in \cite{DBLP:conf/cpaior/AmadiniGM13}. Indeed, also in this case the two best 
portfolios approaches are 3S and SATzilla while the other approaches have lower performances. 
It is worth noticing that we also tried different variants of default WEKA classifiers, obtained by tuning 
parameters, by using meta-classifiers and by performing features selection; despite this changes we did not observed very 
significant differences. Moreover, even in these experiments there is a strong anti-correlation between AST and PSI.
Differently from \cite{DBLP:conf/cpaior/AmadiniGM13}, where the peak performances were reached by relatively small portfolios 
(about 6-8 solvers on a total of 16 solvers), here the addition of a solver almost always increases the percentage of solved instances. 
%
%

Finally, we remark that also in this case the empirical results clearly indicate that a portfolio approach 
could be far better than a single-solver approach: considering for instance the peak performances of 3S on Benchmark B, 
we are able solve up to 25.12\% instances more than the Single Best Solver (SBS) MinisatID, while the maximum gap w.r.t.~the 
Virtual Best Solver is 7.61\%. In particular, 3S is able to close the 89.55\% of the gap between SBS and VBS.

\subsection{Features preprocessing}

We conclude this section with some considerations about features preprocessing. 
In this work we tried to collect a set of features as large and general as possible, obtaining a number of features 
that is more than triple of that one of Mistral.
Obviously, not all of these features are equally significant. 
For example, although in principle MiniZinc allows to use float variables and constraints, none of the 
considered instances contain such constructs. Moreover, having considered only CSPs, all the features related to 
optimization (e.g. solve goal or objective function features) have assumed the same default value. 
Following what is usually done by the majority of current approaches, we then decided to remove all the constant 
features from our features set. In addition, we have scaled all the values in the range [-1, 1]. In this way
we ended up with a reduced set of 114 features on which we conducted our experiments.

\begin{figure}[t]
\centering
\includegraphics[width=0.45\textwidth,trim=1.5cm 14.5cm 1cm 1.5cm, clip]{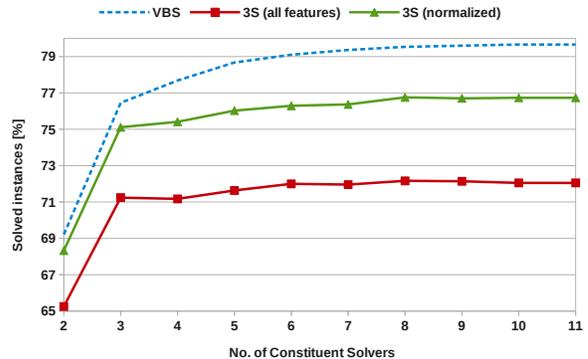}
\caption{PSI comparison of 3S using normalized and not normalized features.}
\label{fig:compare_3S}
\end{figure}

In \cite{citeulike:10395539} the authors show that by using suitable evaluation functions it is possible to 
perform a feature filtering that, on one hand, drastically reduces the feature number and, on the other hand, 
also provides performances gains.
Therefore, we tried to apply different features selection techniques on Benchmark B for all 
the off-the-shelf approaches by exploiting and tuning (using forward, backward and bidirectional search) 
a number of WEKA algorithms, namely: \emph{BestFirst} (a greedy hillclimbing algorithm), 
\emph{GeneticSearch} (based on Bayes Network learning), \emph{GreedyStepwise} (that uses a greedy search), \emph{InfoGainAttributeEval} 
and \emph{Ranker} (for evaluating and ranking the attributes).
Unfortunately, we have not seen significant improvements: all the performance gains were 
always below 1\%.
\footnote{We also tried to apply the filtering techniques of \cite{citeulike:10395539} 
exploiting ISAC code; unfortunately, this proved to be too time consuming (filtering of a single 
fold took more than a day of computation).}
However, it is worth noting that merely removing constant features and scaling them in a given range could lead to 
major enhancements with a small computational effort.
Consider for instance Figure \ref{fig:compare_3S} that shows a comparison between the performances 
obtained by 3S on Benchmark B, by using normalized and not normalized features.
As it can be seen, the difference is considerable: 
the performance gap that can be obtained by normalizing the features ranges from a minimum of 3.08\% 
to a maximum of 4.68\%. 

\section{Conclusions and Extensions}
\label{sec:conclusion}
In this work we presented a framework that is able to extract an 
extensive set of features from both satisfaction and optimization problems defined 
in possibly different modeling languages: MiniZinc, FlatZinc or XCSP.
The main components of the framework are \XCSPZinc, a compiler for converting CSP instances 
from XCSP to MiniZinc, and specially \tool, a flexible and extendible tool for extracting a set of 155 features 
from a MiniZinc model.
We deem that our work could serve as a prototype for the creation of a modern constraint solver 
adopting a portfolio approach. 

Thanks to the features extracted by \tool, it should be pretty straightforward to build a CSP solver 
capable of exploiting a portfolio of different solvers: the only requirement is 
that these solvers must support MiniZinc format. As a future work, we are planning to assemble such 
a solver, possibly enrolling it to a MiniZinc Challenge.

The set of features we proposed was tested using the best performing portfolio techniques evaluated 
in \cite{DBLP:conf/cpaior/AmadiniGM13}. A comparison with all the other existing techniques 
is out of the scope of this paper; nevertheless, we are still interested in testing further portfolio 
approaches (also coming from different domains, e.g. 
Answer Set Programming~\cite{DBLP:conf/lpnmr/GebserKKSSZ11}).

Another future direction concerns the improvement of the quality of the features.
On the one hand, our framework is flexible enough to allow without great effort 
the addition of new and more sophisticated features (SATzilla 
for instance uses a local search algorithm to compute some dynamic features). 
On the other hand, predictions accuracy could be significantly improved by 
appropriate feature filtering techniques.
We noticed that the feature significance depend on the portfolio approach adopted, hence 
we can not provide the subset of the most significant features that works well for all the 
portfolio approaches. Thanks to \tool\ it is however possible to focus on a particular approach and 
devise different filtering techniques for 
improving its performance.


One of the most promising extension of our work is to take into account also Constraint 
Optimization Problems (COPs). In fact, the MiniZinc syntax of COPs is very similar to CSPs:
the only significant difference concerns the solve goal that, in case of optimization, 
defines the integer expression that need to be minimized or maximized.
Since \tool\ is currently already able to process MiniZinc COPs, in the future we are 
planning to investigate if and how a portfolio approach can be effective for solving 
combinatorial optimization problems.
It is our opinion that not all the best portfolio technique practices developed 
for satisfaction problems could be equally good for the optimization field.
For instance, assuming that COPs in average require more time than CSPs to reach the optimal solution, 
an approach like 3S (which first executes short runs of different solvers for the 10\% of the time 
limit) may not be very effective since few COPs could be solved in just 10\% of the allowed time. 
Finally, in order to evaluate a COP solver new metrics should be considered. In fact, often in real 
world it is better to get a good solution in a short time rather than consume too much time to find the optimal value.
Starting from this assumption, it may be reasonable to give each solver a reward proportional to the distance between 
the solution found and the optimal one.

\bibliographystyle{abbrv}
\bibliography{biblio}

\begin{thebibliography}{10}

\bibitem{DBLP:conf/cpaior/AmadiniGM13}
R.~Amadini, M.~Gabbrielli, and J.~Mauro.
\newblock {An Empirical Evaluation of Portfolios Approaches for Solving CSPs}.
\newblock In {\em CPAIOR}, 2013.

\bibitem{Arbelaez09onlineheuristic}
A.~Arbelaez, Y.~Hamadi, and M.~Sebag.
\newblock Online heuristic selection in constraint programming.
\newblock In {\em SoCS}, 2009.

\bibitem{DBLP:conf/ictai/ArbelaezHS10}
A.~Arbelaez, Y.~Hamadi, and M.~Sebag.
\newblock {Continuous Search in Constraint Programming}.
\newblock In {\em ICTAI}, 2010.

\bibitem{ArlotCelisse2010}
S.~Arlot and A.~Celisse.
\newblock A survey of cross-validation procedures for model selection.
\newblock {\em Statistics Surveys}, 4:40--79, 2010.

\bibitem{AHJ+-12-3}
G.~Audemard, B.~Hoessen, S.~Jabbour, J.-M. Lagniez, and C.~Piette.
\newblock {PeneLoPe, a Parallel Clause-Freezer Solver}.
\newblock In {\em SAT Challenge 2012}, pages 43--44, 2012.

\bibitem{fzn_spec}
R.~Becket.
\newblock {Specification of FlatZinc - Version 1.6}.
\newblock \url{http://www.minizinc.org/downloads/doc-1.6/flatzinc-spec.pdf}.

\bibitem{Beldiceanu:2007:GCC:1232658.1232664}
N.~Beldiceanu, M.~Carlsson, S.~Demassey, and T.~Petit.
\newblock {Global Constraint Catalogue: Past, Present and Future}.
\newblock {\em Constraints}, 12(1):21--62, 2007.

\bibitem{snnap}
M.~Collautti, Y.~Malitsky, and B.~O'Sullivan.
\newblock {SNNAP: Solver-based Nearest Neighbor for Algorithm Portfolios}.
\newblock In {\em ECML}, 2013.

\bibitem{CSC09}
{Third International CSP Solver Competition 2008}.
\newblock \url{http://www.cril.univ-artois.fr/CPAI09/}.

\bibitem{DBLP:conf/lpnmr/GebserKKSSZ11}
M.~Gebser, R.~Kaminski, B.~Kaufmann, T.~Schaub, M.~T. Schneider, and S.~Ziller.
\newblock {A Portfolio Solver for Answer Set Programming: Preliminary Report}.
\newblock In {\em LPNMR}, pages 352--357, 2011.

\bibitem{gecode_fzn}
{GECODE flatzinc}.
\newblock \url{http://www.gecode.org/flatzinc.html}.

\bibitem{DBLP:journals/ai/GomesS01}
C.~P. Gomes and B.~Selman.
\newblock Algorithm portfolios.
\newblock {\em Artif. Intell.}, 126(1-2):43--62, 2001.

\bibitem{Hall_theweka}
M.~Hall, E.~Frank, G.~Holmes, B.~Pfahringer, P.~Reutemann, and I.~H. Witten.
\newblock {The WEKA data mining software: an update}.
\newblock {\em SIGKDD Explor. Newsl.}, 11(1), Nov. 2009.

\bibitem{Hamadi09manysat:a}
Y.~Hamadi, S.~Jabbour, and L.~Sais.
\newblock {ManySAT: a Parallel SAT Solver}.
\newblock {\em JSAT}, 6(4):245--262, 2009.

\bibitem{DBLP:journals/corr/abs-1211-0906}
F.~Hutter, L.~Xu, H.~H. Hoos, and K.~Leyton-Brown.
\newblock {Algorithm Runtime Prediction: The State of the Art}.
\newblock {\em CoRR}, 2012.

\bibitem{DBLP:conf/cp/KadiogluMSSS11}
S.~Kadioglu, Y.~Malitsky, A.~Sabharwal, H.~Samulowitz, and M.~Sellmann.
\newblock {Algorithm Selection and Scheduling}.
\newblock In {\em CP}, 2011.

\bibitem{DBLP:conf/ecai/KadiogluMST10}
S.~Kadioglu, Y.~Malitsky, M.~Sellmann, and K.~Tierney.
\newblock {ISAC - Instance-Specific Algorithm Configuration}.
\newblock In {\em ECAI}, 2010.

\bibitem{KMMO11}
Z.~Kiziltan, L.~Mandrioli, J.~Mauro, and B.~O'Sullivan.
\newblock {A classification-based approach to managing a solver portfolio for
  CSPs}.
\newblock In {\em AICS}, 2011.

\bibitem{kotthoff_llama_2013}
L.~Kotthoff.
\newblock {LLAMA: Leveraging Learning to Automatically Manage Algorithms}.
\newblock {\em CoRR}, 2013.

\bibitem{citeulike:10395539}
C.~Kroer and Y.~Malitsky.
\newblock {Feature filtering for instance-specific algorithm configuration}.
\newblock In {\em ICTAI}, 2011.

\bibitem{DBLP:conf/cp/Leyton-BrownNS02}
K.~Leyton-Brown, E.~Nudelman, and Y.~Shoham.
\newblock {Learning the Empirical Hardness of Optimization Problems: The Case
  of Combinatorial Auctions}.
\newblock In {\em CP}, 2002.

\bibitem{DBLP:journals/ai/Mackworth77}
A.~K. Mackworth.
\newblock {Consistency in Networks of Relations}.
\newblock {\em Artif. Intell.}, 8(1):99--118, 1977.

\bibitem{DBLP:conf/ijcai/MalitskySSS13}
Y.~Malitsky, A.~Sabharwal, H.~Samulowitz, and M.~Sellmann.
\newblock Algorithm portfolios based on cost-sensitive hierarchical clustering.
\newblock In {\em IJCAI}, 2013.

\bibitem{DBLP:conf/cpaior/MalitskyS12}
Y.~Malitsky and M.~Sellmann.
\newblock {Instance-Specific Algorithm Configuration as a Method for
  Non-Model-Based Portfolio Generation}.
\newblock In {\em CPAIOR}, 2012.

\bibitem{minisatid}
{KRR Software: MinisatID}.
\newblock \url{http://dtai.cs.kuleuven.be/krr/software/minisatid}.

\bibitem{mistral_old}
{Mistral}.
\newblock \url{http://www.4c.ucc.ie/~ehebrard/mistral/doxygen/html/main.html}.

\bibitem{Mitchell97}
T.~M. Mitchell.
\newblock {\em Machine learning}.
\newblock McGraw Hill series in computer science. McGraw-Hill, 1997.

\bibitem{DBLP:conf/cilc/MoraraMG11}
M.~Morara, J.~Mauro, and M.~Gabbrielli.
\newblock {Solving XCSP problems by using Gecode}.
\newblock In {\em CILC}, 2011.

\bibitem{mznc2012}
{MiniZinc Challenge}.
\newblock \url{http://www.minizinc.org/challenge2012/results2012.html}.

\bibitem{mzn2fzn}
N.~Nethercote.
\newblock {Converting MiniZinc to FlatZinc}.
\newblock \url{http://www.minizinc.org/downloads/doc-1.6/mzn2fzn.pdf}.

\bibitem{Nethercote07minizinc:towards}
N.~Nethercote, P.~J. Stuckey, R.~Becket, S.~Brand, G.~J. Duck, and G.~Tack.
\newblock {MiniZinc: Towards a Standard CP Modelling Language}.
\newblock In {\em CP}, 2007.

\bibitem{cphydra}
E.~O’Mahony, E.~Hebrard, A.~Holland, C.~Nugent, and B.~O’Sullivan.
\newblock Using case-based reasoning in an algorithm portfolio for constraint
  solving.
\newblock {\em AICS 08}, 2009.

\bibitem{ppfolio_web}
O.~Roussel.
\newblock {ppfolio}.
\newblock \url{http://www.cril.univ-artois.fr/~roussel/ppfolio/}.

\bibitem{DBLP:conf/sat/SamulowitzRSS13}
H.~Samulowitz, C.~Reddy, A.~Sabharwal, and M.~Sellmann.
\newblock Snappy: A simple algorithm portfolio.
\newblock In {\em SAT}, pages 422--428, 2013.

\bibitem{source_code}
xcsp2mzn and mzn2feat source code.
\newblock \url{http://www.cs.unibo.it/~amadini/sac_2014.zip}.

\bibitem{DBLP:journals/constraints/StuckeyBF10}
P.~J. Stuckey, R.~Becket, and J.~Fischer.
\newblock {Philosophy of the MiniZinc challenge}.
\newblock {\em Constraints}, 15(3):307--316, 2010.

\bibitem{DBLP:conf/sat/XuHHL12}
L.~Xu, F.~Hutter, H.~Hoos, and K.~Leyton-Brown.
\newblock {Evaluating Component Solver Contributions to Portfolio-Based
  Algorithm Selectors}.
\newblock In {\em SAT}, 2012.

\bibitem{DBLP:conf/cp/XuHHL07}
L.~Xu, F.~Hutter, H.~H. Hoos, and K.~Leyton-Brown.
\newblock {SATzilla-07: The Design and Analysis of an Algorithm Portfolio for
  SAT}.
\newblock In {\em CP}, 2007.

\end{thebibliography}
\end{document}